\documentstyle[ijcai93,named]{article} 

\newcommand{\krterm}[1]{{\sf #1}}
\newcommand{\lingform}[1]{{\it #1}}
\newcommand{\term}[1]{{\it #1}}
\newcommand{\code}[1]{{\tt #1}}
\newcommand{\acronym}[1]{{\sc #1}}
\newcommand{\scare}[1]{`#1'}

\newcommand{\IDAS}{{\sc idas}}

\newcommand{\PENMAN}{{\sc penman}}

\newcommand{\JOYCE}{{\sc joyce}}
\newcommand{\Melchuk}{Mel'\u{c}uk}
\newcommand{\FUF}{{\sc fuf}}

\newcommand{\SPOKESMAN}{{\sc spokesman}}
\newcommand{\MUMBLE}{{\sc mumble}}
\newcommand{\SPL}{{\sc spl}}
\newcommand{\SERC}{{\sc serc}}

\newcommand{\eg}{e.g.}

\title{Has a Consensus NL Generation Architecture Appeared, and is it
Psycholinguistically Plausible?}
\author{Ehud Reiter\thanks{Most of this work was done while the author was at
the University of Edinburgh, Department of Artificial Intelligence.
The Edinburgh work was supported by SERC grant GR/F/36750.}\\
CoGenTex, Inc\\
840 Hanshaw Rd\\
Ithaca, NY 14850 USA\\
email: {\tt ehud@cogentex.com}}

\begin{document}

\bibliographystyle{named}

\maketitle

\begin{abstract}
I survey some recent applications-oriented NL generation systems, and claim
that despite very different theoretical backgrounds, these systems
have a remarkably similar architecture in terms of the modules
they divide the generation process into, the computations these modules
perform, and the way the modules interact with each other.
I also compare
this \scare{consensus architecture} among applied NLG systems
with psycholinguistic knowledge about how humans speak, and argue
that at least some aspects of the consensus architecture seem to be in
agreement with what is known about human language production,
despite the fact that psycholinguistic plausibility was not in general
a goal of the developers of the surveyed systems.
\end{abstract}

%

\section{Introduction}

In this paper I survey some recently-developed NL generation systems
that (a) cover the complete generation process and (b) are designed
to be used by application programs, as well as (or even instead of) making
some theoretical point.  I claim that despite their widely differing
theoretical backgrounds, the surveyed systems are similar in terms of
the modules they divide the generation process into, the way the
modules interact with each other, and (at least in some cases) the kinds
of computations each individual module performs.  In other words,
despite different theoretical claims, there is a remarkable level of
similarity in how these systems \scare{really work};
that is, a {\em de facto} \scare{consensus architecture} seems to be emerging
for how applied NLG systems should generate
text.  The existence of such agreement among the surveyed systems
is especially surprising because in some cases the theoretical backgrounds
of the systems examined argue {\em against} some
aspects of the consensus architecture.

I also compare the consensus architecture to psycholinguistic knowledge
about language generation in human speakers.  Such a comparison is often
difficult to make, because of the many gaps in our current knowledge about how
humans speak.  Nevertheless, I argue that as far as
such a comparison can be made, the specific design decisions embodied in the
consensus architecture seem to often be more or less
in accord with current knowledge of human language generation.  This is again
perhaps somewhat surprising, since psycholinguistic plausibility was not
in general a goal of the developers of the examined systems.  Perhaps
(being very speculative) this indicates that there is some connection between
the engineering considerations that underlie the design decisions made in
the consensus architecture, and the maximize-performance-in-the-real-world
criteria that drove the evolutionary processes that created the human
language processor.  If (a big if!) there is some truth to this hypothesis,
then studying the engineering issues involved in building
applied systems may lead to insights about the way the human language system
works.

\section{The Systems Surveyed}

The analysis presented here is based on a survey of generation systems that:
\begin{enumerate}
\item Were written (or at least substantially extended) since the late 1980s.
This excludes early systems such as Davey's \acronym{proteus} or
Jacobs's \acronym{king}.
\item Are complete systems that start from an intention, a query, or some
data that needs to be communicated, and produce actual sentences as output.
This rules out systems that only implement part of the generation process,
such as Cawsey's \acronym{edge} system (discourse planning)
or my own \acronym{fn} (noun-phrase construction).
\item Were motivated, at least to some degree, by the desire to interface to
application programs.  This excludes
systems that were primarily intended to be computational explorations of a
particular linguistic theory, such as Patten's \acronym{slang},
or computational
models of observed linguistic behavior, such as Hovy's \acronym{pauline}.
\item Are well enough known that I could easily obtain information about them.
\end{enumerate}

In short, the idea was to survey recent systems that looked at the entire
generation problem, and that were motivated by applications and engineering
considerations as well as linguistic theory.
The systems examined were:\footnote{The selection rules are of course
not completely well defined, which means there was inevitably some
arbitrariness when I used them to select particular systems to include in
the survey. I encourage any reader
who believes that I have unfairly omitted a system to contact me, so that
this system can be included in future versions of the survey.}

\begin{description}
\item[FUF] \cite{elhadad:thesis}:
Developed at Columbia University and used in several projects
there, including \acronym{comet} and \acronym{advisor ii}; I will use
the term \scare{\FUF{}} in this paper to refer to both \FUF{}
itself and the various related systems at Columbia.
Several other universities have also recently begun to use \FUF{} in their
research.  \FUF{}  is based on Kay's functional unification formalism
\cite{kay:fug}.
\item[IDAS] \cite{idas:applied-acl}:
Developed at Edinburgh University,
\IDAS{} was a prototype online documentation system for users of complex
machinery.
From a theoretical perspective, \IDAS{}'s main objective was to show that
a single representation and reasoning system can be used for both domain
and linguistic knowledge \cite{idas:acl}.
\item[JOYCE] \cite{joyce:anlp}:
Developed at Odyssey Research Associates, \JOYCE{} is
taken as a representative of several NL generation systems produced by
ORA and CoGenTex, including
\acronym{gossip}, \acronym{fog}, and \acronym{lfs}.  These systems are
all aimed at commercial or government
applications (in \JOYCE{}'s case, producing summaries
of software designs), and are all based on \Melchuk{}'s Meaning-Text theory
\cite{melchuk:book88}.
\item[PENMAN] \cite{penman-user-guide}:
Under development at ISI since the early 1980's,
\PENMAN{} has been used in several demonstration systems.  As usual, I will
use \scare{\PENMAN} to refer to both \PENMAN{} itself and the systems
that were built around it.  \PENMAN{}'s
theoretical basis is systemic linguistics \cite{halliday:book}
and rhetorical-structure theory.
\item[SPOKESMAN] \cite{spokesman:tr}:
\SPOKESMAN{} was developed at BBN for various applications, and has some
of the same design goals as McDonald's \MUMBLE{} system \cite{mcdonald:rnlp},
including in particular the desire to build a system that at least in
some respects is psycholinguistically plausible.  \SPOKESMAN{} uses
Tree-Adjoining Grammars \cite{joshi:tag-nlg} for syntactic processing.
\end{description}
All of the examined systems produce English, and they also are mostly aimed at
producing technical texts (instead of, say, novels or newspaper articles);
it would be interesting to examine systems aimed at other languages or
other types of applications, and see if this caused any architectural
differences.

\section{An Overview of the Consensus Architecture}

As can be seen, the chosen systems have widely different theoretical bases.
It is therefore quite interesting that they all seem to have ended up with
broadly similar
architectures, in that they break up the generation process into a similar
set of modules, and they all use a \term{pipeline architecture} to connect the
modules; i.e., the modules are linearly ordered, and information flows from
each module to its successor in the pipeline, with no feedback from later
modules to earlier modules.
The actual modules possessed by the systems (discussed in more detail
in Section~\ref{details}, as is the pipeline architecture) are:

\begin{description}
\item[Content Determination:]
This maps the initial input of the generation system (\eg{}, a query to be
answered, or an intention to be satisfied) onto a semantic form, possibly
annotated with rhetorical (\eg{}, RST) relations.
\item[Sentence Planning:]
Many names have been used for this process; here I use one suggested
by Rambow and Korelsky \shortcite{joyce:anlp}.
The basic goal is to map conceptual structures onto linguistic ones:
this includes generating referring expressions, choosing content words and
(abstract) grammatical relationships, and grouping information into
clauses and sentences.
\item[Surface Generation:]
I use this term in a fairly narrow sense here, to mean a module that
takes as input an abstract specification of information to be communicated by
syntax and function words, and produces as output a surface form that
communicates this information (\eg{}, maps \code{:speechact imperative}
into an English sentence that lacks a surface subject).
All of the examined systems had separate sentence-planning and
surface-generation modules,
and the various intermediate forms used to pass information
between these modules conveyed similar kinds of information.
\item[Morphology:]
Most of the systems have a fairly simple morphological component, presumably
since English morphology is quite simple.
\item[Formatting:]
\IDAS{}, \JOYCE{}, and \PENMAN{} also contain mechanisms for
formatting (in the \LaTeX{} sense) their output, and/or adding hypertext
annotations to enable users to click on portions of the generated text.
\end{description}

\section{A More Detailed Examination of the Architecture}
\label{details}

This section describes the consensus architecture in more detail, with
particular emphasis on some of the design decisions embodied in it that
more theoretically motivated researchers have disagreed with.  It furthermore
examines the plausibility of these decisions from a psycholinguistic
perspective, and argues that in many respects they agree with what is known
about how humans generate text.

\subsection{Modularized Pipeline Architecture}
\label{modularized}

The consensus architecture divides the generation process into multiple
modules, with information flowing in a \scare{pipeline} fashion from one
module to the next.
By pipeline, I mean that the modules are arranged in a linear order,
and each module receives information only from its
predecessor (and the various linguistic and domain knowledge bases), and
sends information only to its successor.
Information does not flow \scare{backwards} from a module to its predecessor,
and global \scare{blackboards}
that all modules can access and modify are not used.
I do {\em not} mean by \scare{pipeline} that generation must be
incremental in the
sense that, say, syntactic processing of the first sentence is done at the
same time as semantic processing of the second; I believe most of the systems
examined could in fact do this, but they have not bothered to do so
(probably because it would not be of much benefit to the applications programs
of interest).

\subsubsection{Design decision: avoid integrated architecture}
\label{ellis-examples}

Many NL generation researchers have argued against
dividing the generation process into modules; perhaps the best-known are
Appelt \shortcite{appelt:book} and Danlos \shortcite{danlos:acl84}.
Others, such as Rubinoff \shortcite{rubinoff:inlgw92},
have accepted modules but have argued that the architecture must allow feedback
between later modules and earlier modules, which argues against the
one-way information flow of the pipeline architecture.

The argument against pipelines and modules is almost always some variant of
\scare{there are linguistic phenomena that can only be properly handled by
looking at constraints from different levels
(intentional, semantic, syntactic, morphological),
and this is difficult to do in a pipeline system.}
To take one fairly random example,
Danlos and Namer \shortcite{danlos:coling88} have
pointed out that since the French masculine and
feminine pronouns \lingform{le} and \lingform{la} are abbreviated to
\lingform{l'} before a word that starts with a vowel, and since in some
cases \lingform{le} and \lingform{la} may be unambiguous references while
\lingform{l'} is not, the referring expression system must have some
knowledge of surface word
order and selected content and function words before it can decide whether
a pronoun is acceptable; this will not be possible if referring expressions are
chosen before syntactic structures are built, as happens in the consensus
architecture.

There is undoubtably some truth to these arguments, but the applications
builder also has to consider the engineering reality that the sorts of systems
proposed by Appelt, Danlos, and Namer are extremely difficult to build from an
engineering perspective.  The engineering argument for modularization is
particularly strong;  Marr has put this very well in
\cite[page 485]{marr:rs76}:
\begin{quote}
Any large computation should be split up and implemented as a collection of
small subparts that are as nearly independent of one another as the overall
task allows.  If a process is not designed in this way a small change in
one place will have consequences in many other places.  This means that the
process as a whole becomes extremely difficult to debug or improve, whether
by a human designer or in the course of natural evolution, because a small
chance to improve one part has to be accompanied by many simultaneous
compensatory changes elsewhere.
\end{quote}
Marr argues that a modularized structure makes
sense both for human engineers and for the evolutionary process that produced
the human brain.  The evidence is indeed strong
that the human brain is highly modularized.  This evidence comes from many
sources (\eg{}, cognitive experiments and PET scans of brain activity), but
I think perhaps the most convincing evidence is from studies of humans
with brain damage.  Such people tend to lose specific abilities, not suffer
overall degradation that applies equally to all abilities.
Ellis and Young \shortcite{ellis-young:book} provide an
excellent summary of such work, and list patients that, for example
\begin{itemize}
\item can produce syntactically correct utterances but can not
organize utterances into coherent wholes, i.e., can perform surface generation
but not content determination.
\item can generate word streams that tell a narrative but are not organized
into sentences, i.e., can perform content determination but not surface
generation.
\item can produce coherent texts organized in grammatical structures, but
have a severely restricted vocabulary; i.e., have impaired lexical choice
(these patients still have conceptual knowledge, they just have problems
lexicalizing it).
\end{itemize}

The main engineering argument for arranging modules into a pipeline
instead of a more complex structure is again simplicity and ease of debugging.
In a one-way pipeline
of N modules there are only N-1 interfaces between modules, while a pipeline
with `two-way' information flow has 2(N-1) interfaces, and a system that
fully connects each module with every other module
will have N(N-1) interfaces.  A system that has a
two-way interface between every possible pair of modules will undoubtably
be able to handle many linguistic phenomena in a more powerful, elegant,
principled, etc, manner
than a system that arranges modules in a simple one-way pipeline;
such a system will also, however, be much more difficult to build
and (especially) debug.

It is easy to argue
that a one-way pipeline is worse at handling some linguistic phenomena
than a richly-connected architecture, but this is not the end of the story
for the system-building engineer; he or she has to balance the cost of the
pipeline being inefficient and/or inelegant at handling some phenomena
against the benefit of the pipeline being a much easier structure to build
and debug.  We have insufficient engineering data at present to make any
well-substantiated claims about
whether the one-way pipeline has the optimal cost/benefit tradeoff or not
(and in any case this will probably depend somewhat on the circumstances
of each application \cite{idas:ijcai}),
but the circumstantial evidence on this question is striking; despite
the fact that so many theoretical papers have argued against pipelines and
very few (if any) have argued for pipelines, every one of the
applications-oriented systems examined in this survey chose to use the
one-way pipeline architecture.

In other words, an applications systems builder can not
look at particular linguistic phenomena in isolation;
he or she must weigh the benefits of \scare{properly} handling these
phenomena against the cost of implementing the proposed architecture.
In the French pronoun case described by Danlos and Namer, for example,
the applications builder might argue
that in the great majority of cases no harm will in fact be done if the
referring-expression generator simply ignores the possibility that pronouns
may be abbreviated to \lingform{l'}, especially given humans' ability
to use context to disambiguate references; and if a situation does arise where
it is absolutely essential that the human reader be able to correctly
disambiguate a reference, then perhaps pronouns should not be used in any
case.  Given this,
and the very high engineering cost of building an integrated architecture of
the sort proposed by Danlos and Namer,
is implementing such an architecture truly the most effective way of using
scarce engineering resources?

Psycholinguistic research on self-monitoring and self-repair (summarized in
\cite[pages 458-299]{levelt:book}) suggests that there is some
feedback in the human language generation system, so the human
language processor is probably more complex than a simple one-way pipeline;
but it may not be much more complex.  To the best of my knowledge, most
of the observed self-repair phenomena could be explained by an architecture
that added a few feedback loops from later stages of the pipeline back
to the initial planner; this would only slightly add to the number of
inter-module interfaces (perhaps N+1 instead of N-1, say), and hence would
have a much lower engineering cost than implementing the fully connected
`every module communicates with every other module' architecture.
Whether the human language
engine is organized as a `pipeline plus a few feedback loops' or
an `every module talks to every other module'
architecture is unknown at this point; hopefully new
psycholinguistic experiments will shed more light on this issue.
I think it would be very interesting, for example,
to test human French speakers on
situations of the sort described by Danlos and Namer, and see what they
actually did in such contexts; I do not believe that such an experiment
has (to date) been performed.

\subsection{Content Determination}
\label{content determination}

Content determination takes the initial input to the generation system,
which may be,
for example, a query to be answered or an intention to be satisfied, and
produces from it a \scare{semantic form}, \scare{conceptual representation},
or \scare{list of propositions},
i.e., a specification of the meaning
content of the output text.  I will in this paper use the term
\term{semantic representation} for this meaning specification.
Roughly speaking, the semantic
representations used by all of the examined systems can be characterized
as some kind of \scare{semantic net} (using the term in its broadest sense,
as in \cite{sowa:book})
where the primitive elements in the net are conceptual instead of linguistic
(\eg{}, domain KB concepts instead of English words).
In some cases the semantic nets also
include discourse and rhetorical relations between portions of the net;
subsequent portions of the generator use these
to generate discourse connectives (\eg{}, \lingform{However}),
control formatting (\eg{}, the use of bulletized lists), etc.

The systems examined use quite different content-determination mechanisms
(i.e., there was no consensus); schemas \cite{mckeown:book} were
the most popular approach.

\subsubsection{Design decision: integrated content determination and
rhetorical planning}

Content determination in the systems examined basically performs two
functions:
\begin{description}
\item[Deep content determination:]
Determine what information should be communicated to the hearer.
\item[Rhetorical planning:]
Organize this information in a rhetorically coherent manner.
\end{description}

Hovy \shortcite{hovy:acl88} has proposed an architecture where these tasks are
performed separately (in particular, the application program performs deep
content determination, while the generation system performs rhetorical
planning).
Among the systems examined, however, Hovy is unique in taking
this approach; the builders of the other systems (including Moore and Paris
\shortcite{moore-paris:acl},
who also worked with \PENMAN{}) apparently believe that
these two processes are so closely related that they should be performed
simultaneously.

I am not aware of any psychological data that
directly address this issue.  However, Hovy's architecture
requires the language-producing agent to completely determine the content
of a paragraph before he/she/it can begin to utter it (since the rhetorical
planner determines what the first sentence is, and it is not called until
deep content determination is completed), and intuitively it seems implausible
to me that human speakers do this;
it also goes against incremental theories of human speech production
\cite[pages 24-27]{levelt:book}.

\subsection{Sentence Planning}

The sentence planner converts the semantic representation, which is specified
in terms of domain entities, into an abstract linguistic representation
that specifies content words and grammatical relationships.  I will use
\Melchuk{}'s term \term{deep syntactic} form for this representation.

All of the systems analyzed possess a deep syntactic representation;
none attempt to go from semantics to surface form in a single step.
\IDAS{} and \PENMAN{} use variants of the same deep syntactic language,
\SPL{} \cite{kasper:darpa-workshop}.
\FUF{} and \JOYCE{} use deep syntactic languages that are based (respectively)
on functional unification and meaning-text theory, but these convey
much the same information as \SPL{}.
\SPOKESMAN{} uses the realization specification language of \MUMBLE{}
\cite{mcdonald:rnlp}
as its deep syntactic representation; I have found it difficult to compare this
language to the others, but McDonald (personal communication) agrees that
it conveys essentially the same information as \SPL{}.

Unfortunately,
while all of the systems possessed a module which converted semantic
representations into deep syntactic ones, each system used a different
name for this module.
In \FUF{} it is the \scare{lexical chooser},
in \IDAS{} it is the \scare{text planner}, in \JOYCE{} it is
the \scare{sentence planner}, in \SPOKESMAN{} it is the
\scare{text structurer}, and in \PENMAN{} it doesn't seem to have a name
at all, \eg{}, Hovy \shortcite{hovy:acl88}
simply refers to \scare{pre-generation text-planning
tasks}.  I use the \JOYCE{} term here because I think it is the least
ambiguous.

The specific tasks performed by the sentence planner include:
\begin{enumerate}
\item Mapping domain concepts and relations into content words and
grammatical relations.
\item Generating referring expressions for individual domain entities.
\item Grouping propositions into clauses and sentences.
\end{enumerate}

Relatively little is said in the papers about clause grouping and
referring-expression generation,
but more information is available on the first task,
mapping domain entities onto linguistic entities.
All the examined systems except perhaps \PENMAN{} use a variant of what
I have elsewhere
called the \scare{structure-mapping} approach \cite{reiter:ci};\footnote{Even
though I have previously argued against structure-mapping because it does not
do a good job of handling lexical preferences \cite{reiter:ci},
I nevertheless ended up using this technique when I moved from my Ph.D research
to the more applications-oriented \IDAS{} project.  Perhaps this is another
example of engineering considerations overriding theoretical arguments.}
I do not know what approach \PENMAN{} uses (the papers are not clear on this).
Structure-mapping is based
on a dictionary that lists the semantic-net equivalents of
\term{linguistic resources} \cite{meteer:ci} such as content words
and grammatical
relationships.  This dictionary might, for example, indicate that
the English word \lingform{sister} is equivalent (in the domain knowledge-base
of interest) to the structure \krterm{Sibling with attribute Sex:Female},
and that the domain relation \krterm{Part-of} can be expressed with
the grammatical possessive, \eg{}, \lingform{the car's engine}.
Given this dictionary, the structure-mapping algorithm iteratively replaces
semantic structures by linguistic ones, until the entire semantic net has been
recoded into a linguistic structure.  There may be several ways of
recoding a semantic representation into a linguistic one, which means
structure-mapping systems have a choice between using the first acceptable
reduction they find, or doing a search for a reduction that maximizes some
optimality criterion (\eg{}, fewest number of words).  The papers I read were
not very clear on this issue, but I believe that while most of the systems
surveyed use the first acceptable reduction found,
\FUF{} in some cases searches for an optimal reduction.

\subsubsection{Design decision:
separation of lexical choice from surface realization}

The consensus architecture clearly separates lexical choice of content
words (done during sentence planning) from syntactic processing
(performed during surface generation).  In other words, it does {\em not}
use an integrated \scare{lexicogrammar}, which systemic theorists in
particular (\eg{}, \cite{matthiessen:chapter})
have argued for, and which is implicit in some
unification-based approaches, such as the semantic head-driven algorithm
\cite{shieber:cl90}.

Despite these theoretical arguments, none of the systems examined used
an integrated lexicogrammar, including unification-based \FUF{} and
systemic-based \PENMAN{}.\footnote{The \PENMAN{} papers do not explicitly
say where lexical choice is performed.  However,
all examples of \PENMAN{} \SPL{} input
that I have seen have essentially had content words already specified, which
suggests that lexical choice is performed before syntactic
processing in \PENMAN{}.}
In contrast, earlier unification-based systems,
such as the tactical component of McKeown's \acronym{text}
system \cite{mckeown:book}, did integrate
lexical and syntactic processing in a single \scare{tactical generator};
also, systemic systems that have been less
driven by application needs than \acronym{penman},
such as \acronym{genesys} \cite{genesys:coling90}, have used
integrated lexicogrammars.

There is psychological evidence that at least some lexical processing is
separated from syntactic processing, \eg{}, the patient mentioned
in Section~\ref{ellis-examples} who
was able to perform content-determination and syntactic generation
but had a very restricted speaking vocabulary.  I think it's also very
suggestive that humans have different learning patterns for content
and function words; the former are \scare{open-class} and easily learned,
while the latter are \scare{closed-class} and people tend to stick to the
ones they learned as children.
There is less evidence on the location of lexical choice in
the psycholinguistic pipeline, and on whether
it is performed in one stage or distributed among several stages.

\subsection{Surface Generation}

\term{Surface generation} has been used to mean many different things in
the literature.  I use it here to refer to the portion of the generation
system that knows how grammatical relationships are actually expressed
in English (or whatever the target language is).
For example, it is the surface generator that knows what
function words and word order relationships are used in English for
imperative, interrogative, and negated sentences; it is the surface
generator that knows which auxiliaries are required for the various
English tenses; and it is the surface generator that knows when
pronominalization is syntactically required (\lingform{John scolded himself},
not \lingform{John scolded John}).

\subsubsection{Design decision: top-down algorithm with (almost?) no
backtracking}

The grammars and grammar representations used by the systems examined are
quite different, but all systems process the grammars with a top-down
algorithm that uses minimal, if any, backtracking.
None of the systems use the semantic head-driven generation
algorithm \cite{shieber:cl90}, although this is probably the single
best-known algorithm for surface generation;
Elhadad \shortcite[chapter 4]{elhadad:thesis}
claims that such an algorithm is only necessary
for systems that attempt to simultaneously perform both lexical choice and
surface generation, which none of the examined systems do.
Perhaps more interestingly, four of the five systems do not allow
backtracking, and the fifth, \FUF{}, allows backtracking but does not seem to
use it much (if at all) during surface generation (backtracking is used in
\FUF{} during sentence planning).
This is interesting, since backtracking is usually regarded as an essential
component of unification-based generation approaches;
it is certainly used in the
semantic-head-driven algorithm, and in the \acronym{text}
generator \cite{mckeown:book}.

From a psycholinguistic perspective, many people have argued that human
language production is incremental
(see the summary in \cite[pages 24-27]{levelt:book}),
which means that of necessity it cannot include much backtracking.
The garden-path phenomena shows that there are
limits to how much syntactic backtracking people people perform during
language understanding.
This evidence is of course suggestive rather than
definitive; it seems likely that there are limitations on how much (if any)
backtracking humans will perform during syntactic
processing (see also the arguments in \cite{mcdonald:rnlp}),
but there is no hard proof of this (as far as I am aware).

\subsection{Morphology and Formatting}

These modules will not be further examined here, mainly because
little information is given in the papers on the details of how morphology
and formatting are implemented.

\section{A Controversial (?) View}

I would like to conclude with a perhaps controversial personal opinion.
There have been many cases where NL generation researchers
(including myself) have claimed that a certain linguistic phenomena is
best handled by a certain architecture.
Even if this is true, however, if it turns out that
adopting this architecture will substantially complicate
the design of the overall generation system,
and that the most common cases of the
phenomena of interest can be adequately handled by adding a few heuristics
to the appropriate stage of a simpler architecture,
then the engineering-oriented NL worker must ask him- or herself
if the benefits of the proposed architecture truly outweigh its costs.
For instance, one cannot simply argue that an integrated architecture
is superior to a pipeline because it is better suited to handling certain
kinds of pronominalization; it is also necessary to evaluate the engineering
cost of shifting to an integrated architecture, and determine if, for example,
better overall performance for the amount of engineering resources available
could be obtained by keeping the general pipeline architecture, and instead
investing some of the engineering resources \scare{saved} by this decision
into building more sophisticated heuristics into the pronominalization module.

In doing so, I believe (and again this is a personal belief that probably
cannot be substantiated by the existing evidence)
that the NL engineer is coming close to the \scare{reasoning}
of the evolutionary process that created the human language system.
Evolution does not care about
elegant declarative formalisms or \scare{proper} (as opposed to \scare{hacky})
handling of special cases; evolution's goal is to maximize performance in
real-world situations, while maintaining an architecture that can be easily
tinkered with by future evolutionary processes.
In short, evolution is an engineer, not a mathematician.\footnote{Gould's
various popular books on evolutionary biology,
such as \cite{gould:book}, give
an excellent feel for evolution as an engineer-cum-hackers;
see also the interesting discussion of language and evolution
in \cite{pinker:book}.}
It is thus
perhaps not surprising if NL generation systems designed to be used in
real-world applications end up with an architecture that seem to bear some
resemblance to the architecture of the
human language processor;\footnote{Of course, the best way to do something on
a machine is often not the best way to do it in nature; \eg{}, birds and
airplanes use different mechanisms to fly.  On the other hand, there does seem
to be a remarkable congruence between effective vision processing strategies
in animals and computers \cite{marr:book}.  One could also argue that since
language (unlike flying) is purely a product of the human mind,
any effective language processor is probably
going to have to share some of the mind's processing strategies.}
and future attempts to build applications-oriented generation systems
may end up giving us real insights into how language processing works in
humans, even if this is not the main purpose of these systems.
Similarly, psycholinguistic knowledge of how the human language generator
works may suggest useful algorithms for NL engineers;
one such case is described in \cite{reiter-dale:coling92}.

Cross-fertilization between psycholinguistics and NL engineering
will only arise, however, if the results of engineering
analyses are reported in the research literature, especially
when they suggest going against some theoretical principle.
Unfortunately, to date the results of such
analyses have all-too-often been regarded more as embarrassments
(since they contradict theory) than as valuable observations,
and hence have not been published.
I would like to conclude this paper by encouraging generation researchers
to regard the results of engineering analyses to be as interesting
and as important to the understanding of language as conventional
linguistic analyses.  After all, as Woods \shortcite{woods:tinlap}
has pointed out, while
descriptive analyses of language can at best tell us {\em what} the brain
does, engineering analyses can potentially offer insights on
{\em why} the brain functions as it does.

\section*{Acknowledgements}

I would like to thank Jean Carletta, Robert Dale, Michael Elhadad,
David McDonald, Richard Kittredge, Tanya Korelsky,
Chris Mellish, Owen Rambow, and Graeme Ritchie
for their very helpful comments on earlier versions of this work.
It goes without saying, of course, that the views represented are my own,
and that any factual errors are entirely my fault.
This research was mostly done while the author was at the University of
Edinburgh, where he was supported by \SERC{} grant GR/F/36750.



\end{document}